\documentclass{article}
\PassOptionsToPackage{numbers,square,sort}{natbib}
\usepackage[dblblindworkshop,final]{neurips_2025}

\usepackage[utf8]{inputenc}
\usepackage[T1]{fontenc}
\usepackage{hyperref}
\usepackage{url}
\usepackage{booktabs}
\usepackage{amsfonts}
\usepackage{nicefrac}
\usepackage{microtype}
\usepackage{xcolor}
\usepackage{graphicx}

\title{Position: AI Will Transform Neuropsychology Through Mental Health Digital Twins for Dynamic Mental Health Care, Especially for ADHD}

\author{
  Neil Natarajan, Sruthi Viswanathan, Xavier Roberts-Gaal, and Michelle Marie Martel \\
}

\begin{document}

\maketitle

\begin{abstract}
Static solutions dont serve a dynamic mind. Thus, we advocate a shift from static mental health diagnostic assessments to continuous, artificial intelligence (AI)-driven assessment. Focusing on Attention-Deficit/Hyperactivity Disorder (ADHD) as a case study, we explore how generative AI has the potential to address current capacity constraints in neuropsychology, potentially enabling more personalized and longitudinal care pathways. In particular, AI can efficiently conduct frequent, low-level experience sampling from patients and facilitate diagnostic reconciliation across care pathways. We envision a future where mental health care benefits from continuous, rich, and patient-centered data sampling to dynamically adapt to individual patient needs and evolving conditions, thereby improving both accessibility and efficacy of treatment. We further propose the use of mental health digital twins (MHDTs) -- continuously updated computational models that capture individual symptom dynamics and trajectories -- as a transformative framework for personalized mental health care. We ground this framework in empirical evidence and map out the research agenda required to refine and operationalize it.
\end{abstract}

\section{Introduction}
Mental health conditions are dynamic; presentation and impact shift with treatment response, life events, and development \citep{Dijk_Mierau_2023}. Attention-Deficit/Hyperactivity Disorder (ADHD) illustrates this dynamism well: it often persists into adulthood and affects roughly 4\% of adults worldwide, yet its symptomatology varies across contexts and over the lifespan \citep{Franke_Michelini_Asherson_Banaschewski_Bilbow_Buitelaar_Cormand_Faraone_Ginsberg_Haavik_etal._2018}. Hyperactivity tends to dominate in preschool years, whereas inattention becomes central from middle childhood onward; DSM subtypes show limited temporal stability \citep{Lahey_Pelham_Loney_Lee_Willcutt_2005,Martel_vonEye_Nigg_2012}. Traditional point-in-time assessments struggle to capture these dynamics \citep{Donovan_Ellis_Cole_Townsend_Cases_2020,Kazda_Bell_Thomas_McGeechan_Sims_Barratt_2021}. We argue that recent advances in generative AI can directly address these constraints and enable patient-centered, longitudinal assessment. Throughout, we use ADHD as a running case to illustrate both the limitations of current practice and the potential of AI-assisted approaches.

This position paper makes three contributions: (i) it explains why episodic neuropsychological assessment is misaligned with dynamic conditions, using ADHD as exemplar; (ii) it proposes using mental health digital twin (MHDT) systems -- a clinician-supervised, multimodal, continuously updated model -- to support longitudinal diagnostic reconciliation in response; and (iii) it outlines a concrete research and operational agenda to validate, govern, and safely deploy such systems.

\section{Limitations of Current Neuropsychological Assessment}\label{sec:problem}
\begin{figure}[t]
\centering
\includegraphics[width=\textwidth]{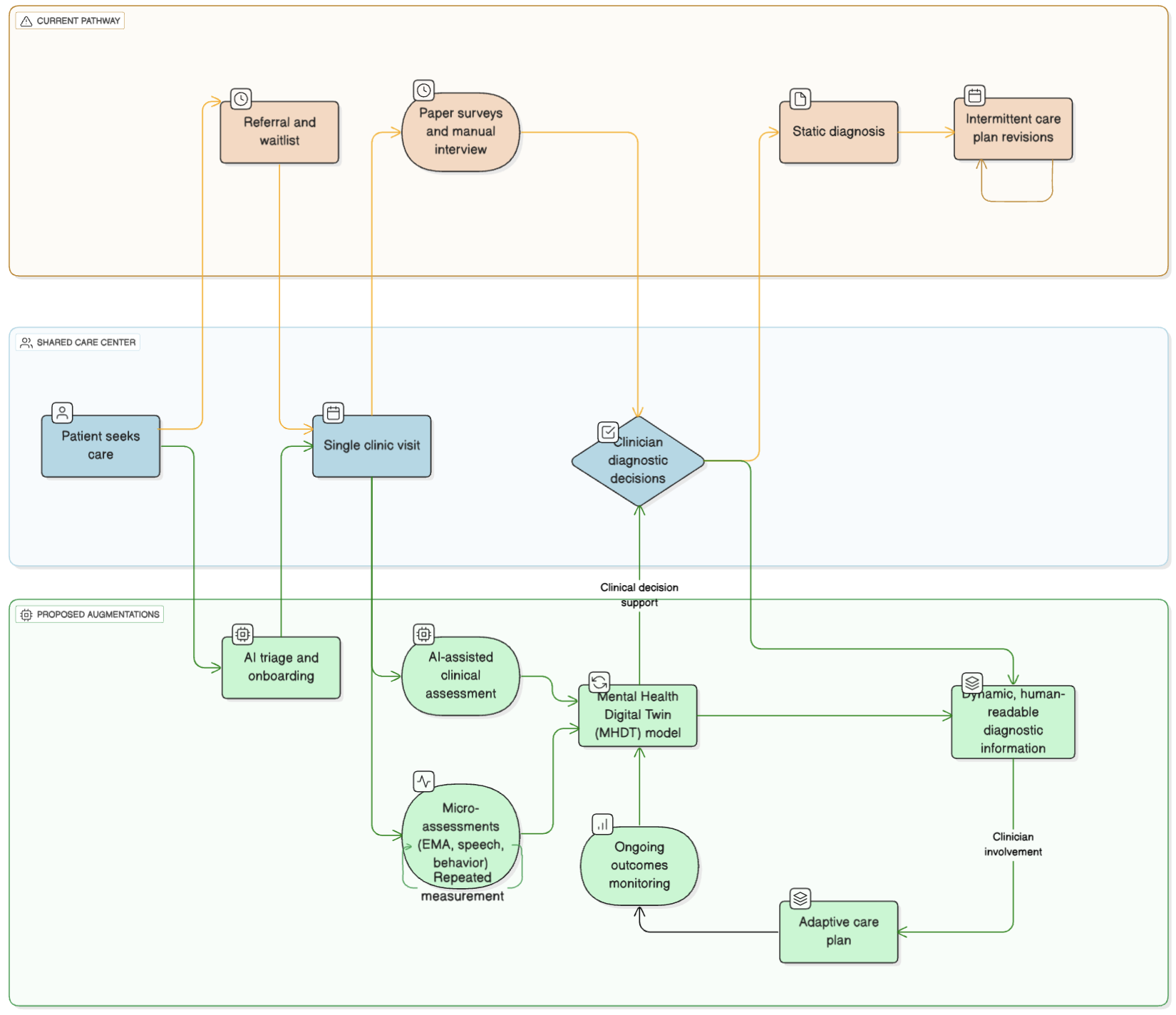}
\caption{Comparison of traditional episodic ADHD diagnostic process versus continuous AI-assisted diagnostic companion approach. The traditional model (top) shows discrete assessment points separated by long intervals, while the continuous model (bottom) demonstrates ongoing data collection, dynamic modeling, and iterative diagnostic refinement.}
\label{fig:diagnostic-processes}
\end{figure}

The evolution of many mental health conditions over time complicates single-occasion diagnostics. For ADHD, adult diagnosis often relies on retrospective reconstruction of childhood onset, frequently without reliable records \citep{Culpepper_Mattingly_2010}. Longitudinal cohort evidence identifies a fluctuant subgroup whose diagnostic status changes with age, consistent with relapsing–remitting patterns or sensitivity to developmental context \citep{Norman_Price_Ahn_Sudre_Sharp_Shaw_2023}. Taken together with developmental shifts in symptom structure and instability of Diagnostic and Statistical Manual of Mental Disorders (DSM) subtypes \citep{Lahey_Pelham_Loney_Lee_Willcutt_2005,Martel_vonEye_Nigg_2012}, this motivates longitudinal, dimensional assessment in place of single-point categorical assignment.

Current neuropsychological workflows typically compress evaluation into episodic interactions over short intervals \citep{Culpepper_Mattingly_2010,Donovan_Ellis_Cole_Townsend_Cases_2020}. Such snapshots miss within-person variation across time and settings, creating risks of misclassification as new information renders prior labels obsolete without systematic reconciliation \citep{Donovan_Ellis_Cole_Townsend_Cases_2020,AlgarinPerneth_PerezRodriguezGarcia_Brito_Gandhi_Bylund_Hargraves_SinghOspina}. In practice, many adult ADHD determinations are made via rating scales and brief clinical encounters \citep{Hechtman_French_Mongia_Cherkasova_2011}, and re-assessment requires new visits, extending delays and costs \citep{Donovan_Ellis_Cole_Townsend_Cases_2020,Culpepper_Mattingly_2010}.

Structural constraints further limit ecological validity: primary care lacks capacity for longitudinal, context-rich evaluations, while specialty clinics face long waitlists \citep{Culpepper_Mattingly_2010,Koziol_Stevens_2012}. Office-bound assessments struggle to capture context effects (home, school, work) and time-varying presentations \citep{Musullulu_2025}. Consequently, transient states can be frozen into durable labels \citep{Lahey_Pelham_Loney_Lee_Willcutt_2005,Martel_vonEye_Nigg_2012}.

\section{Dynamic Approaches to Diagnosis}\label{sec:solution}
Ecological momentary assessment (EMA), a method for collecting real-time data on symptoms, behaviors, and experiences in participants' natural environments through repeated brief surveys delivered via smartphones or other devices, has emerged as a promising approach to continuous monitoring of mental health symptoms \citep{Miguelez-Fernandez_deLeon_Baltasar-Tello_Peñuelas-Calvo_Barrigon_Capdevila_Delgado-Gómez_Baca-García_Carballo_2018,Kennedy_Molina_Pedersen_2024}. However, purely EMA-based monitoring has practical limitations: it relies on self-report, creates burden and attrition risk, and may itself alter symptom reports (measurement reactivity) \citep{Miguelez-Fernandez_deLeon_Baltasar-Tello_Peñuelas-Calvo_Barrigon_Capdevila_Delgado-Gómez_Baca-García_Carballo_2018,Kennedy_Molina_Pedersen_2024}. EMA also under-samples nonlinguistic and physiologic correlates relevant to attention and affective regulation \citep{Miguelez-Fernandez_deLeon_Baltasar-Tello_Peñuelas-Calvo_Barrigon_Capdevila_Delgado-Gómez_Baca-García_Carballo_2018}.

Going beyond EMA, AI systems can coordinate adaptive prompts and fuse EMA with other modalities through ongoing, clinician-supervised conversational loops. Unlike human clinicians, these systems can interact continuously, enabling longitudinal data collection before, during, and after the traditional diagnostic process \citep{Bongurala_Save_Virmani_2024}. Furthermore, generative AI systems have demonstrated capacity for real-time, personalized interactions that adapt to user behavior, increasing patient engagement and allowing for on-demand interactions based on the most recent changes in symptom presentation \citep{Orru_Melis_Sartori_2025}. Indeed, these AI systems could prompt patients for interactions across various environments, gathering data difficult for clinicians to obtain directly, enabling more holistic diagnostic understanding. These capabilities present an opportunity to build high-touch diagnostic systems capable of capturing information over time, improving on retrospective analyses.

This concept may be pushed even further through the development of mental health digital twins (MHDTs), i.e., continuously updated, individualized computational models of a patient's mental states and likely trajectories \citep{Spitzer_Dattner_Zilcha-Mano_2023}. While the MHDT remains largely theoretical \citep{Tu_Schaekermann_Palepu_Saab_Freyberg_Tanno_Wang_Li_Amin_Cheng_etal._2025}, it provides a blueprint for future research and development. These AI models would track and predict patient presentation by integrating data from patient interactions, behavioral patterns, and physiological responses, moving beyond single-point diagnoses to continuous, personalized care \citep{Bongurala_Save_Virmani_2024}. By using these tools for longitudinal diagnosis in addition to care, clinicians may glean insights into evolving conditions beyond the scope of what point assessments, even ones assisted by AI systems, can capture \citep{Spitzer_Dattner_Zilcha-Mano_2023}.

An ADHD-focused diagnostic companion built on MHDT principles would integrate: (i) structured conversational history-taking aligned to DSM constructs and comorbidity screening \citep{Tu_Schaekermann_Palepu_Saab_Freyberg_Tanno_Wang_Li_Amin_Cheng_etal._2025,Jiang_Shen_Lai_Qi_Zheng_Yao_Wang_Pan_2024}; (ii) brief at-home cognitive and performance tasks (e.g., continuous performance test [CPT] variants) for longitudinal tracking of attention and inhibition \citep{Cedergren_Ostlund_AsbergJohnels_Billstedt_Johnson_2022}; (iii) short speech probes for prosodic and linguistic markers associated with ADHD \citep{vonPolier_Ahlers_Volkening_Langner_Patil_Eickhoff_Helmhold_Krautz_Langner_2025,Barrios_Poznyak_LeeSamson_Rafi_Gabay_Cafiero_Debbane_2025}; and (iv) optional wearable-derived behavioral and neurophysiologic signals when appropriate and consented \citep{Merritt_Zak_2024}. Clinician-in-the-loop triage and reporting remain central to ensure fitness-to-practice and guard against over-pathologizing normal variability.

The MHDT would explicitly model symptom dynamics, secondary features (e.g., speech prosody and language markers, home-based cognitive performance tasks, or wearable signals), and context effects in latent space. This allows for forward simulation and counterfactual reasoning, enabling clinicians to test hypotheses about the impact of environmental structure or medication adjustments under uncertainty bounds. Evidence supporting each feature family includes prosodic and linguistic markers \citep{vonPolier_Ahlers_Volkening_Langner_Patil_Eickhoff_Helmhold_Krautz_Langner_2025,Barrios_Poznyak_LeeSamson_Rafi_Gabay_Cafiero_Debbane_2025}, at-home cognitive tasks \citep{Cedergren_Ostlund_AsbergJohnels_Billstedt_Johnson_2022}, wearable neurophysiology in mood prediction \citep{Merritt_Zak_2024}, and known EMA dynamics such as measurement reactivity \citep{Miguelez-Fernandez_deLeon_Baltasar-Tello_Peñuelas-Calvo_Barrigon_Capdevila_Delgado-Gómez_Baca-García_Carballo_2018,Kennedy_Molina_Pedersen_2024}, under-sampling of nonlinguistic and physiologic correlates \citep{Miguelez-Fernandez_deLeon_Baltasar-Tello_Peñuelas-Calvo_Barrigon_Capdevila_Delgado-Gómez_Baca-García_Carballo_2018}, or predictable cyclic modulators (e.g., menstrual cycle) \citep{Roberts_Eisenlohr-Moul_Martel_2018}. In combination, these digital twins could transform the diagnostic reconciliation process by continuously analyzing new information and updating diagnoses, moving beyond repeated reconciliation to a fundamentally evolving understanding of patient conditions.

\section{A Mental Health Digital Twin Research Agenda}\label{sec:agenda}
Several challenges stand in the way of making MHDT a reality. Data quality and generalizability remain open questions, as signal-behavior relationships may vary across demographics and contexts \citep{Adler_Stamatis_Meyerhoff_Mohr_Wang_Aranovich_Sen_Choudhury_2024}. Continuous monitoring introduces privacy, consent, and governance complexities beyond traditional records \citep{Tan_Sumner_Wang_WenjunYip_2024,Weiner_Dankwa-Mullan_Nelson_Hassanpour_2025}. Bias and fairness concerns must be proactively addressed, given evidence of differential performance across subgroups and potential racial bias in model-mediated recommendations \citep{Bouguettaya_Stuart_Aboujaoude_2025,Timmons_Duong_Fiallo_Lee_Vo_Ahle_Comer_Brewer_Frazier_Chaspari_2023}. There is risk of diagnostic drift or over-pathologizing variability if update thresholds are not conservative and well-calibrated \citep{Kazda_Bell_Thomas_McGeechan_Sims_Barratt_2021}. Finally, integrating with existing workflows while avoiding alert fatigue, ensuring safety guardrails, and maintaining human accountability are essential \citep{Sutton_Pincock_Baumgart_Sadowski_Fedorak_Kroeker_2020,Weiner_Dankwa-Mullan_Nelson_Hassanpour_2025}.

\textbf{Research milestones:}
\begin{itemize}
\item \textit{Validated AI-delivered intake probes.} Build and evaluate AI applications for history-taking and standardized tasks (modalities as in Section~\ref{sec:solution}), with longitudinal scoring to support clinician diagnosis. Validate via uplift-style trials comparing diagnostic accuracy and cost to clinician-only baselines \citep{Jaskowski_Jaroszewicz}. Demonstrate non-inferiority or superiority while reducing assessment burden.
\item \textit{Multimodal MHDT modeling.} Develop calibrated fusion of EMA, speech, behavioral, and physiologic data with quantified uncertainty; detect signal and population drift; and maintain provenance across model updates \citep{Spitzer_Dattner_Zilcha-Mano_2023}.
\item \textit{Diagnostic reconciliation policies.} Specify conservative posterior-thresholds for label changes, minimum evidence windows, rate limits on update frequency, and confirmatory clinician reviews; quantify false-update risk and clinician workload under different policies \citep{AlgarinPerneth_PerezRodriguezGarcia_Brito_Gandhi_Bylund_Hargraves_SinghOspina,Sutton_Pincock_Baumgart_Sadowski_Fedorak_Kroeker_2020}.
\end{itemize}

\textbf{Operationalization priorities:}
\begin{itemize}
\item \textit{Informed, revocable consent and patient control.} Granular, ongoing consent with clear opt-in/out and patient portals to view, pause, or delete data where permissible \citep{Tan_Sumner_Wang_WenjunYip_2024,Weiner_Dankwa-Mullan_Nelson_Hassanpour_2025}.
\item \textit{Inclusive calibration and drift safeguards.} Use conservative thresholds, confirmatory assessments before label changes, and rate-limit updates \citep{Kazda_Bell_Thomas_McGeechan_Sims_Barratt_2021}. Calibrate to account for known EMA dynamics \citep{Kennedy_Molina_Pedersen_2024,Roberts_Eisenlohr-Moul_Martel_2018}, so updates reflect durable change rather than transient state fluctuations.
\item \textit{Data governance, protection, and response.} Data minimization, on-device processing when feasible, and encryption in transit/at rest \citep{Tan_Sumner_Wang_WenjunYip_2024}; role-restricted access, minimal retention, and lineage/versioning \citep{Weiner_Dankwa-Mullan_Nelson_Hassanpour_2025}; and transparent incident response aligned with GDPR/HIPAA requirements \citep{Shah_2023}.
\item \textit{Transparency and interpretability.} Clinician-readable explanations, uncertainty estimates, and data/model provenance; user interface (UI) that supports shared decision-making via interpretable longitudinal visualizations \citep{Sutton_Pincock_Baumgart_Sadowski_Fedorak_Kroeker_2020,Weiner_Dankwa-Mullan_Nelson_Hassanpour_2025}.
\item \textit{Human oversight and governance.} Explicit override pathways, audit trails, and multidisciplinary governance to review audits, approve model changes, and manage deprecation \citep{AlgarinPerneth_PerezRodriguezGarcia_Brito_Gandhi_Bylund_Hargraves_SinghOspina,Weiner_Dankwa-Mullan_Nelson_Hassanpour_2025,Sutton_Pincock_Baumgart_Sadowski_Fedorak_Kroeker_2020}.
\item \textit{Bias and fairness management.} Pre-specified subgroup audits, equity-drift monitoring, stakeholder reviews, and mitigations for differential performance and racial bias \citep{Adler_Stamatis_Meyerhoff_Mohr_Wang_Aranovich_Sen_Choudhury_2024,Bouguettaya_Stuart_Aboujaoude_2025,Timmons_Duong_Fiallo_Lee_Vo_Ahle_Comer_Brewer_Frazier_Chaspari_2023}.
\item \textit{Safety and escalation.} Escalation protocols for risk signals integrated to minimize alert fatigue while ensuring timely response \citep{Sutton_Pincock_Baumgart_Sadowski_Fedorak_Kroeker_2020,Weiner_Dankwa-Mullan_Nelson_Hassanpour_2025}.
\end{itemize}

\section{A Vision for Personalized Mental Health Models Beyond Diagnostic Categories}\label{sec:expansion}

Looking further into the future, the MHDT framework may ultimately transcend traditional diagnostic categories altogether. The fundamental purpose of psychiatric diagnosis is to predict treatment response and guide clinical decision-making. However, categorical diagnoses like ADHD, depression, or anxiety often fail to capture heterogeneity and comorbidity \citep{Norman_Price_Ahn_Sudre_Sharp_Shaw_2023}. Evidence that categorical subtypes are unstable across development \citep{Lahey_Pelham_Loney_Lee_Willcutt_2005} and that symptom loadings shift with age \citep{Martel_vonEye_Nigg_2012} further supports dimensional, individualized modeling. A general ``p'' factor of psychopathology may explain cross-diagnostic covariance, motivating transdiagnostic modeling \citep{Caspi_Houts_Belsky_Goldman-Mellor_Harrington_Israel_Meier_Ramrakha_Shalev_Poulton_etal._2014,Adam_2023}.

MHDTs offer an alternative: individualized models that directly predict treatment responses, functional outcomes, and support needs. These models may capture each patient's constellation of symptoms, contexts, biological markers, and treatment history, avoiding brittle intermediary labels. For example, rather than ``ADHD with comorbid anxiety,'' a profile would indicate likely response to behavioral interventions, expected benefit from stimulant medication under certain environmental conditions, and the need for accommodations.

This paradigm shift acknowledges the dimensional nature of mental health and better accounts for developmental trajectories and context. It directly optimizes for treatment outcomes and functional improvement rather than labels. It is also evidence-generating: routine prospective measurement coupled with causal inference can enrich both clinical care and research.

However, this vision faces substantial challenges beyond those discussed in Section~\ref{sec:problem}. Medical systems and insurance frameworks are organized around diagnostic categories \citep{Yu_Gorgone_2025}; and regulatory approval often presumes categorical indications \citep{VanNorman_2016}. Clinicians are trained—and incentivized—to think in categorical terms using the DSM and the International Classification of Diseases (ICD) \citep{First_Westen_2007} and routine intake practices \citep{Nakash_Nagar_Kanat-Maymon_2015}. Patients, too, can experience diagnostic labels as meaningful identities, with both stigma and persistence of impressions even after retraction \citep{Garand_Lingler_Conner_Dew_2009,Mickelberg_Walker_Ecker_Fay_2024}. Transitioning to a fully personalized, model-based approach will require both technical and institutional change.

The future of mental wellbeing lies in companions (AI or otherwise) that adapt to evolving psychological landscapes. Static, episodic diagnostic models are poorly matched to dynamic conditions and capacity constraints. A mental health digital twin can combine conversational history-taking, repeatable at-home probes, and selective passive sensing to support diagnostic reconciliation over time under clinician oversight. We propose a research agenda and a roadmap toward MHDTs that are interpretable, auditable, safe, and clinically useful.

Realizing this vision requires rigorous validation across populations, careful attention to privacy and governance, explicit bias and fairness auditing, human-centred design, and seamless integration into clinical workflows. If these milestones are met, continuous AI-supported diagnostics could reduce misclassification, shorten time-to-assessment, reduce cost, and enable more precise, measurement-based care. The ambition is not to replace clinicians, but to equip them with longitudinal, high-fidelity evidence that makes diagnosis and treatment more accurate and responsive to change. The ultimate promise of this position is a future where no mind is left without timely, precise, accessible, and dynamic care.

\newpage
\bibliographystyle{plainnat}
\bibliography{ref}

\end{document}